\documentclass[10pt,twocolumn,letterpaper]{article}

\usepackage{cvpr}
\usepackage{times}
\usepackage{epsfig}
\usepackage{graphicx}
\usepackage{amsmath}
\usepackage{amssymb}
\usepackage{color}

\usepackage{gensymb}
\usepackage{float}
\usepackage{multirow}
\usepackage{diagbox, eqparbox, hhline}
\usepackage{booktabs,tabularx,graphicx}

\usepackage[pagebackref=true,breaklinks=true,letterpaper=true,colorlinks,bookmarks=false]{hyperref}

\newcommand{\tref}[1]{Table~\ref{#1}}

\newcommand{\fref}[1]{Fig.~\ref{#1}}
\newcommand{\sref}[1]{Sec.~\ref{#1}}
\cvprfinalcopy 

\def\cvprPaperID{1113} 

\ifcvprfinal\fi
\begin{document}
	
	\title{EPINET: A Fully-Convolutional Neural Network \\Using Epipolar Geometry for Depth from Light Field Images}
	
	\author{Changha Shin$^1$\hspace{7mm}Hae-Gon Jeon$^2$\hspace{6mm}Youngjin Yoon$^2$\hspace{5mm}In So Kweon$^2$\hspace{8mm}Seon Joo Kim$^1$\\
		$^1$Yonsei University \hspace{10mm} $^2$KAIST\\
		{\tt\scriptsize changhashin@yonsei.ac.kr}\hspace{2mm}{\tt\scriptsize earboll@kaist.ac.kr}\hspace{2mm}{\tt\scriptsize jeromeyoon@kaist.ac.kr}\hspace{2mm}{\tt\scriptsize iskweon@kaist.ac.kr}
		\hspace{2mm}{\tt\scriptsize seonjookim@yonsei.ac.kr}
	}
	
	\maketitle
	\thispagestyle{empty}
	
	\begin{abstract}
		{Light field cameras capture both the spatial and the angular properties of light rays in space.
			Due to its property, one can compute the depth from light fields in uncontrolled lighting environments, which is a big advantage over active sensing devices. 
			Depth computed from light fields can be used for many applications including 3D modelling and refocusing. 
			However, light field images from hand-held cameras have very narrow baselines with noise, making the depth estimation difficult. 
			Many approaches have been proposed to overcome these limitations for the light field depth estimation, but there is a clear trade-off between the accuracy and the speed in these methods. 
			In this paper, we introduce a fast and accurate light field depth estimation method based on a fully-convolutional neural network. 
			Our network is designed by considering the light field geometry and we also overcome the lack of training data by proposing light field specific data augmentation methods. 
			We achieved the top rank in the HCI 4D Light Field Benchmark on most metrics, and we
			also demonstrate the effectiveness of the proposed method on real-world light-field images.
		}
		\\ \vspace{-0.10cm} 
	\end{abstract}
	
	\section{Introduction}
	Light field cameras collect and record light coming from different directions. As one of the most advanced techni-ques introduced in the area of computational photography, the new hand-held light field camera design has a broad impact on photography as it changes how images are captured and enables the users to alter the point of view or focal plane after the shooting.  
	
	Since the introduction of the camera array system~\cite{Wilburn05}, many approaches for making compact and hand-held light field cameras have been proposed like the lenslet-based cameras~\cite{Ng05,Lytro,LytroIllum,Raytrix}, which utilize a micro-lens array placed in front of the imaging sensor.
	Images captured from the lenslet-based cameras can be converted into multi-view images with a slightly different view point via geometric calibration processes~\cite{Bok17,Dansereau13}.
	Thanks to this special camera structure, light field cameras can be used to estimate the depth of a scene in uncontrolled environments.
	This is one of the key advantages of the light field cameras over active sensing devices~\cite{Kinect,Tango}, which require controlled illumination and are therefore limited to the indoor use. 
	\begin{figure}[!t]
		\centering
		\hspace{-0.35cm}
		\includegraphics[page=1,width=1.0\linewidth,trim={0cm 0cm 2.7cm 0cm},clip]{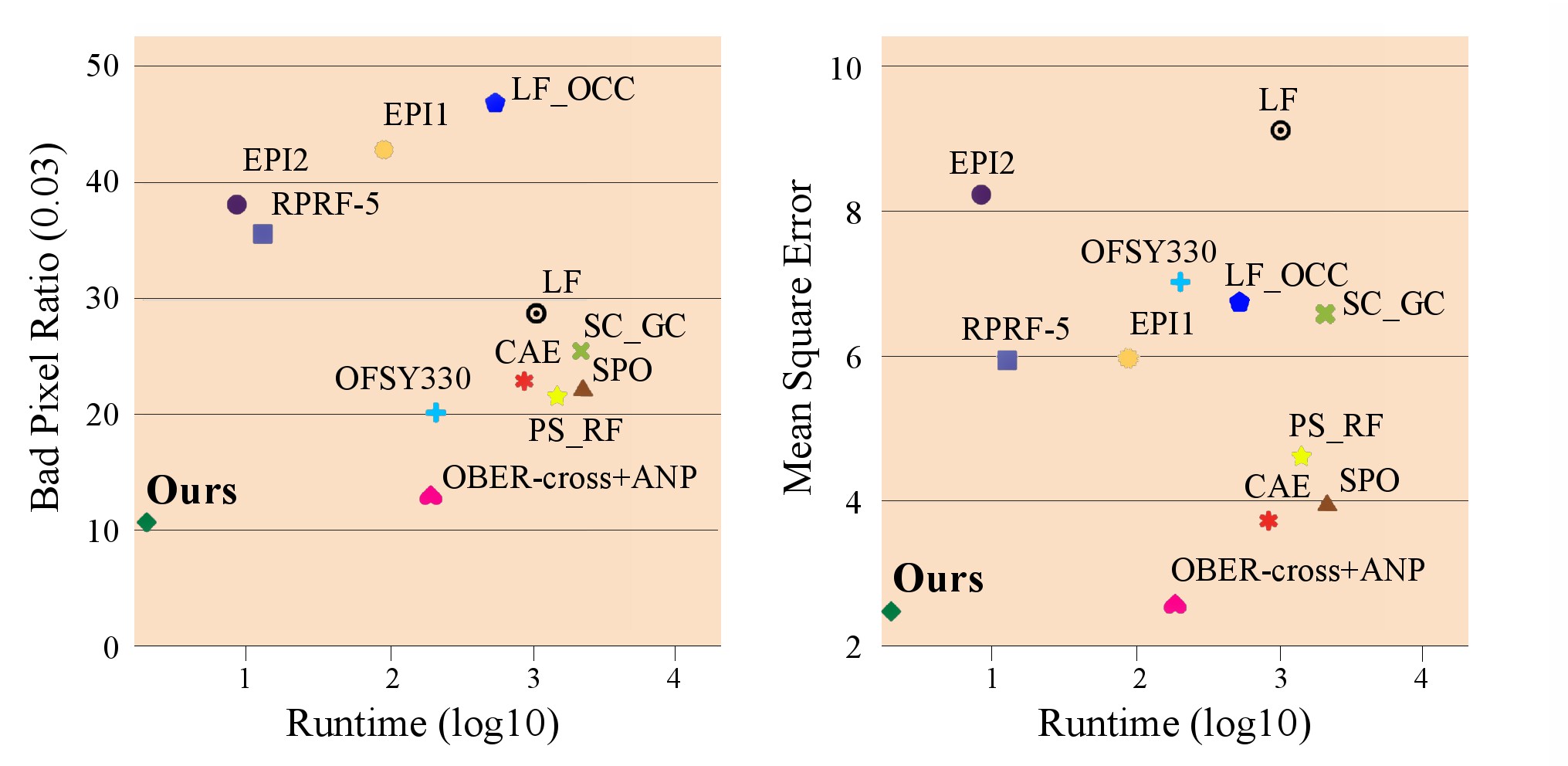}
		\caption{Comparison of the accuracy and the run time of light field depth estimation algorithms. 
	Our method achieved the top rank in the HCI 4D Light Field Benchmark, both in accuracy and speed. 
	Our method is 85 times faster than OBER-cross+ANP, which is a close second in the accuracy ranking. }
		\label{Fig:f1}

\end{figure}		

	On the other hand, the hand-held light field cameras have their own limitation.
	Due to their structure, the baseline between sub-aperture images is very narrow and there exists a trade-off between the spatial and the angular resolution within the restricted image sensor resolution. 
	Various approaches~\cite{Tao13,Jeon15,Heber14,Yu13,Wang16,Zhang16} have been introduced to overcome these limitations and
	acquire accurate depth maps.
	These methods achieve good performances close to other passive sensor-based depth estimation methods such as the stereo matching, but are less practical due to their heavy computational burden.
	Although several fast depth estimation methods~\cite{Wanner13,huang2017robust} have been proposed, they lose the accuracy to gain the speed.

	In this paper, we introduce a deep learning-based approach for the light field depth estimation that achieves both accurate results and fast speed.   
	Using a convolutional neural network, we estimate accurate depth maps with sub-pixel accuracy in seconds.
	We achieved the top rank in HCI 4D Light Field Benchmark\footnote{\url{http://hci-lightfield.iwr.uni-heidelberg.de}}\cite{Honauer16} on most quality assessment metrics including the bad pixel ratio, the mean square error, and the runtime as shown in~\fref{Fig:f1}.
	
	In our deep network design, we create four separate, yet identical processing streams for four angular directions (horizontal, vertical, left and right diagonal) of sub-aperture images and combine them at a later stage.
	With this architecture, the network is constrained to first produce meaningful representations of the four directions of the sub-aperture images independently.
	These representations are later combined to produce higher level representation for the depth estimation. 
	
	One problem of applying deep learning for light field depth estimation is the lack of data.
	Publicly available light field datasets do not contain enough data to train a deep network. 
	To deal with this problem of data insufficiency, we additionally propose a data augmentation method specific for the light field imaging.
	We augment the data through scaling, center view change, rotation, transpose, and color that are suitable for light field images.
	Our data augmentation plays a significant role in increasing the trainability of the network and the accuracy of the depth estimation.
	\vspace{0.2cm}
	
	\section{Related Work}
	The related works can be divided in two categories: depth from a light field image using optimization approaches and learning-based approaches. 
	\vspace{0.3cm} \\ 
	\noindent{\bf Optimization-based methods.} 
	The most representative method for the depth estimation using light field images is the use of the epipolar plane images (EPIs), which consist of 2D slices angular and spatial directions~\cite{Levoy96light,Gortler96}. 
	As the EPI consists of lines with various slopes, the intrinsic dimension is much lower than its original dimension.
	This makes image processing and optimization tractable for the depth estimation.
	Wanner and Goldluecke~\cite{Wanner13} used a structured tensor to compute the slopes in EPIs, and refined initial disparity maps using a fast total variation denoising filter. 
	Zhang~\etal~\cite{Zhang16} also used the EPIs to find the matching lines and proposed a spinning parallelogram operator to remove the influence of occlusion on the depth estimation.
	
	Another approach is to exploit both defocus and correspondence cues.
	Defocus cues perform better in repeating textures and noise, and correspondence cues are robust in bright points and features. 
	Tao~\etal~\cite{Tao13} first proposed a depth estimation that combines defocus and correspondence cues.
	This approach was later improved by adding shading-based refinement technique in~\cite{Tao16} and a regularization with an occlusion term in~\cite{Wang16}.
	Williem and Park~\cite{Williem16} proposed a method robust to noise and occlusion.
	It is equipped with a novel data cost using an angular entropy metric and adaptive defocus responses.
	
	Many other methods have been proposed to improve the depth estimation from light field image.
	Heber and Pock~\cite{Heber14} proposed a global matching term which formulates a low rank minimization on the stack of sub-aperture images.
	Jeon~\etal~\cite{Jeon15} adopted a multi-view stereo matching with a phase-based sub-pixel shift.
	The multi-view stereo-based approach enabled the metric 3D reconstruction from a real-world light field image. 
	
	These conventional optimization based methods have an unavoidable trade-off between the computational time and the performance. 
	In this paper, we adopt a convolutional neural network framework to gain both the speed and the accuracy.
	
	\vspace{0.3cm} 
	\noindent{\bf Learning based methods.} 
	Recently, machine learning techniques have been applied to a variety of light field imaging applications such as super-resolution~\cite{Yoon15,yoon:2017}, novel view generation~\cite{Kalantari16}, single image to a light field image conversion~\cite{srinivasan2017learning}, and material recognition~\cite{Wang16material}.
	
	For the depth estimation, Johannsen~\etal~\cite{johannsen2016sparse} presented a technique which uses EPI patches to compose a dictionary with a corresponding known disparity. 
	This method yielded better results on multi-layered scenes.
	Heber~\etal~\cite{heber2017neural} proposed an end-to-end deep network architecture consisting of an encoding and a decoding part. 
	Heber and Pock~\cite{Heber16} proposed a combination of a CNN and a variational optimization. They trained a CNN to predict EPI line orientations, and formulated a global optimization with a higher-order regularization to refine the network predictions.
	
	There are still issues in the aforementioned learning based methods. 
	    Those methods only consider one directional epipolar geometry of light field images in designing the network \cite{Heber16, heber2017neural}, resulting in low reliability of depth predictions.
		We overcome this problem via a multi-stream network which encodes each epipolar image separately to improve the depth prediction. 
		Because each epipolar image has its own unique geometric characteristics, we separate epipolar images into multiple parts to make the deep network to take advantage of the characteristics.
		Another issue is that the insufficient training data limit the discriminative power of the learned model and lead to over-fitting.
        In this work, we propose novel data augmentation techniques for light-field images that lead to good results without the over-fitting issue.
	
	
	\begin{figure*}[t]
		\centering
		\hspace{0em}
		\includegraphics[page=1,width=0.91\linewidth,trim={0cm 2.6cm 0cm 0.2cm},clip]{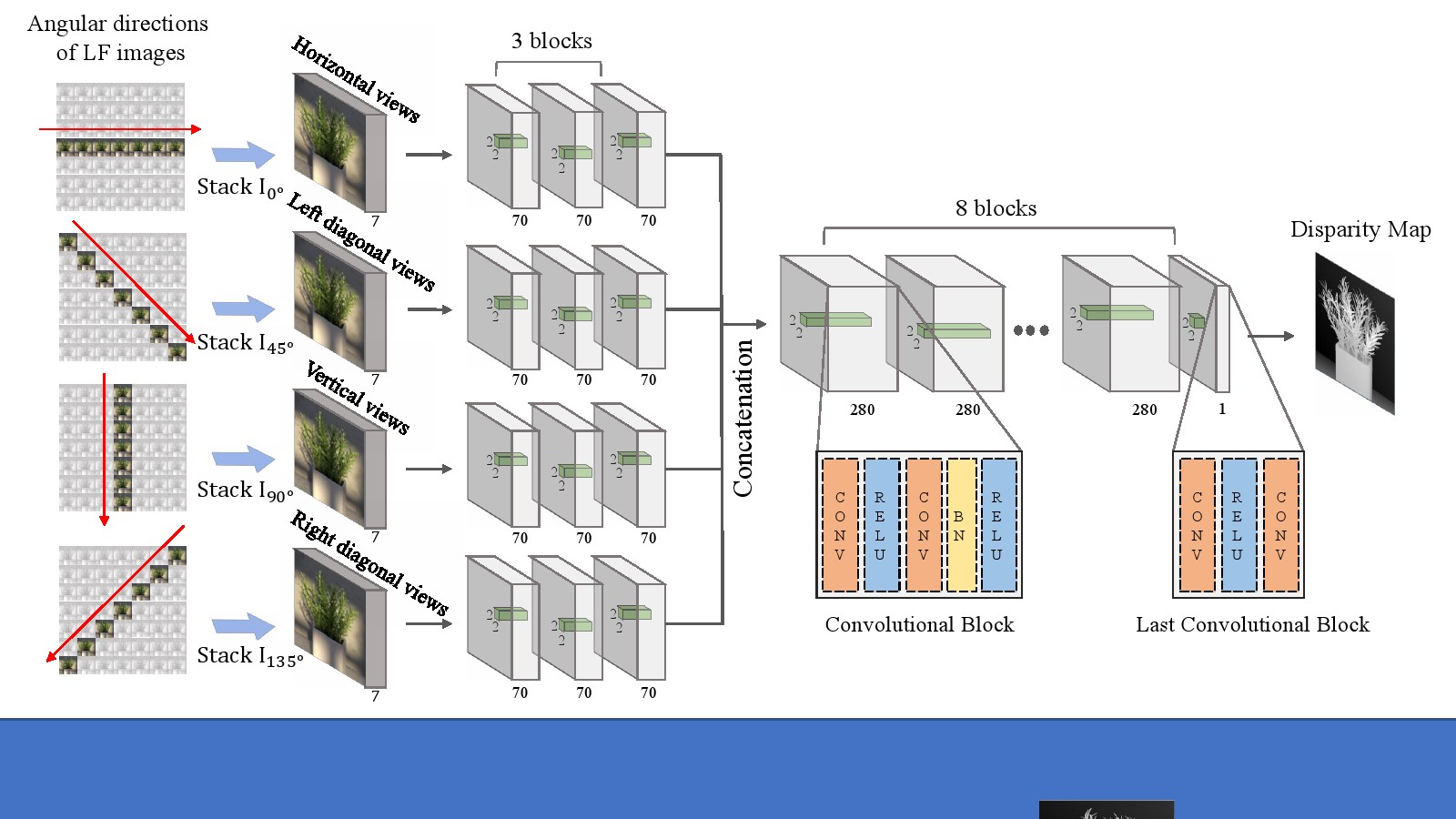}
		\caption{EPINET: Our light field depth estimating architecture.}
		\label{Fig:f2}
	\end{figure*}
	\section{Methodology}
	
	\subsection{Epipoloar Geometry of Light Field Images}
	\label{ch:LF}
	With the insights from previous works, we design an end-to-end neural network architecture for the depth from a light-field image exploiting the characteristics of the light field geometry.
	Since the light field image has many angular resolutions in the vertical and the horizontal directions, the amount of data is much bigger than that of a stereo camera.
	When using all the viewpoints of the light field images as an input data, despite the accurate light field depth results, the computation speed is several hundred times slower than the stereo depth estimation algorithm. To solve this problem, several papers proposed algorithms that only use horizontal or crosshair viewpoints of light field images ~\cite{Wanner13} ~\cite{johannsen2016sparse} ~\cite{Zhang16} ~\cite{SAG17cvpr}. Similarly, we propose a depth estimation pipeline by first reducing the number of images to be used for the computation by exploiting the light field characteristics between the angular directions of viewpoints.
	
	The 4D light field image is represented as $L(x, y, u, v)$, where $(x, y)$ is the spatial resolution and $(u, v)$ is the angular resolution. The relationship between the center and the other viewpoints of light field image can be expressed as follows: 
	\begin{equation}
	L(x, y, 0, 0) = L(x + d(x, y)*u, y + d(x, y)*v, u, v),
	\end{equation}
	where $d(x, y)$ is the disparity of the pixel $(x,y)$ in the center viewpoint from its corresponding pixel in its adjacent viewpoint. 
	
	For an angular direction $\theta$ $(\tan \theta = v/u)$, we can reformulate the relationship as follows:
	\begin{align}
	L (x, y, 0, 0) & = L(x + d(x,y)*u, y + d(x,y) * u\tan\theta, \nonumber \\
	& \hspace{15pt} u, u\tan\theta )
	\label{eq:2}
	\end{align}
	
	However, the viewpoint index is an integer, so there are no corresponding viewpoints when $\tan\theta$ is non-integer.
	Therefore, we select images in the direction of four viewpoint angles $\theta$: 0, 45, 90, 135 degrees assuming that the light field images have $(2N+1)\times (2N+1)$ angular resolution.
	\vspace{0.1cm}
	
	\subsection{Network Design}
\begin{figure}[t]
		\centering
		\includegraphics[page=1,width=0.5\textwidth,trim={0cm 10cm 5cm 0cm},clip]{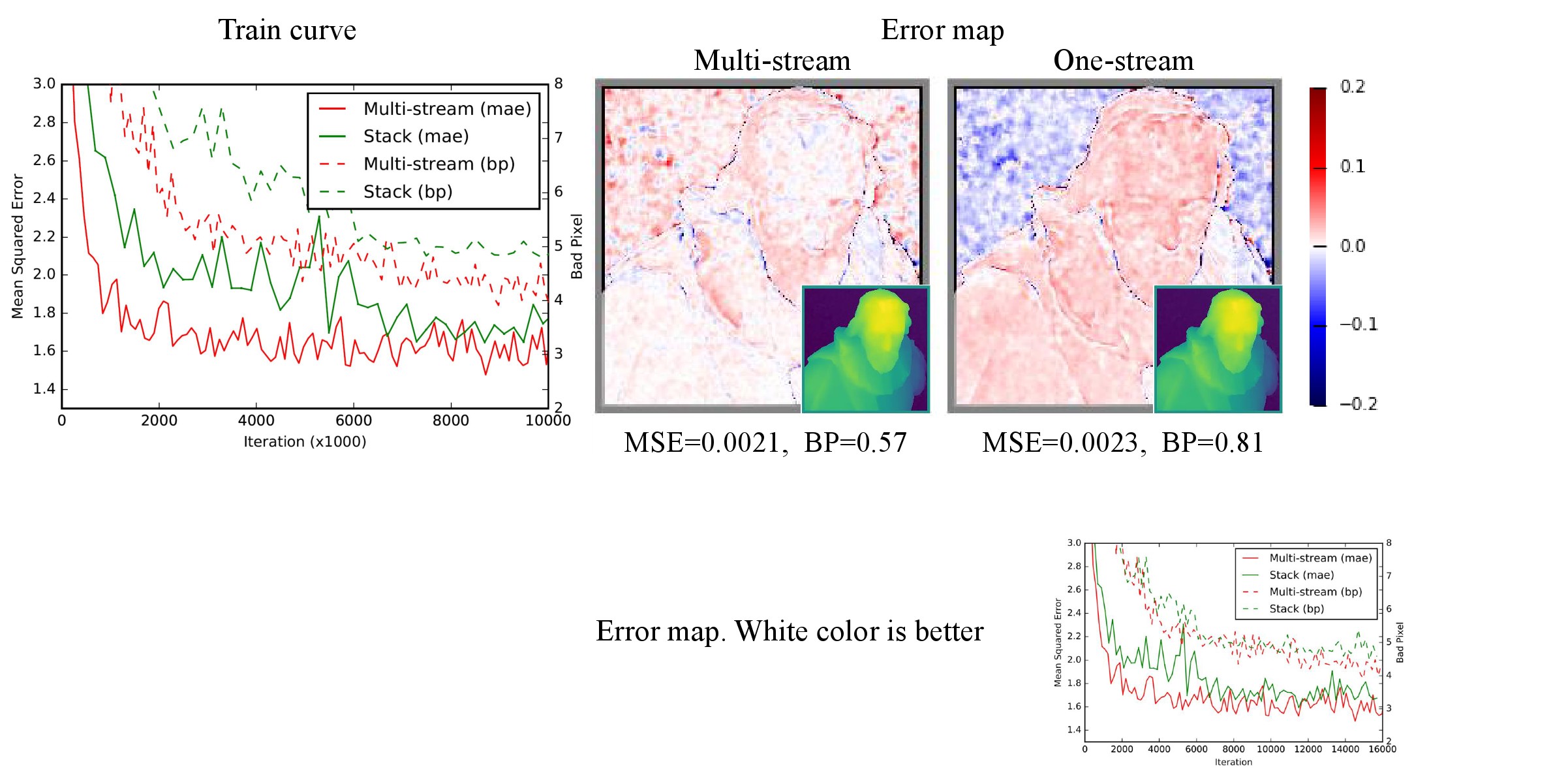}
		\caption{Multi-stream vs one-stream: Test error during training (left), and error map at the 10M iteration (right). In the error map, white color represents lower errors.}
		\label{Fig:multistream}
		\vspace{-0cm}
	\end{figure}

	\noindent{\bf Multi-stream network.}
	As shown in \fref{Fig:f2}, we construct a multi-stream networks for four viewpoints with consistent baselines based on \sref{ch:LF}: horizontal, vertical, left and right diagonal directions. 
		Similar to the conventional optical flow estimation and stereo matching approaches~\cite{dosovitskiy2015flownet,Luo16}, we encode each image stack separately in the beginning of our network. In order to show the effectiveness of the multi-stream architecture, we quantitatively compare our multi-stream network to a single stream network. 
		As shown in~\fref{Fig:multistream}, the reconstruction error using the proposed method is about 10 percent lower, even when using the same number of parameters as the one-stream network.
		With this architecture, the network is constrained to first produce meaningful representations of the four viewpoints.

	The multi-stream part consists of three fully convolutional blocks.
	Since fully convolutional networks are known to be effective architecture for pixel-wise dense prediction~\cite{long2015fully}, we define a basic block with a sequence of fully convolutional layers: 'Conv-ReLU-Conv-BN-ReLU' to measure per-pixel disparity in a local patch.
	To handle the small baseline of light-field images, we use small 2x2 kernel with stride 1 to measure a small disparity value ($\pm$4 pixels).
	\begin{table}[t]
		\centering
		\caption{The effect of the number of viewpoints on performance.}
		\label{tab:stream}
		\begin{tabular}{|c|@{\hspace{0.0cm}}c|@{\hspace{0.0cm}}c|@{\hspace{0.0cm}}c|}
			\hline 
			\multicolumn{1}{|c|}{} & \multicolumn{1}{c|}{1-stream} & \multicolumn{1}{c|}{2-streams} & \multicolumn{1}{c|}{4-streams} \\ \hline \hline
			\small{Input Views}  & 
			\noindent\hrulefill\par
			\noindent\makebox[.09\textwidth][c]{
				\begin{minipage}{.09\textwidth}
					\centering
					\rule{0pt}{0ex}   
					\includegraphics[page=1,width=1\textwidth,trim={0cm 0.8cm 16.5cm 1cm},clip]{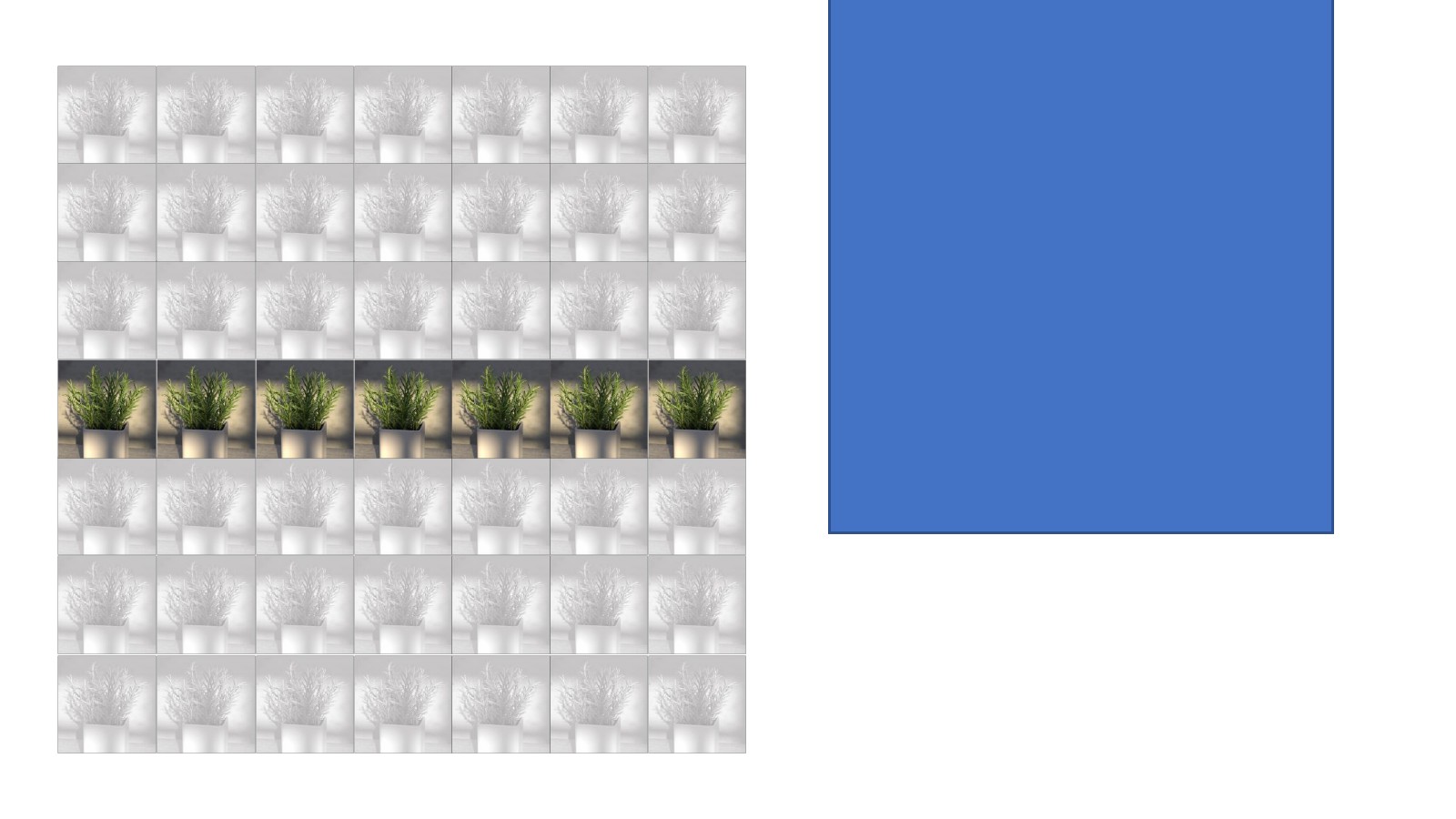} 
			\end{minipage}}
			& 
			\noindent\makebox[.09\textwidth][c]{
				\begin{minipage}{.09\textwidth}
					\centering
					\hspace{0.5cm}
					\rule{0pt}{0ex} 
					\includegraphics[page=1,width=1\textwidth,trim={0cm 0.8cm 16.5cm 1cm},clip]{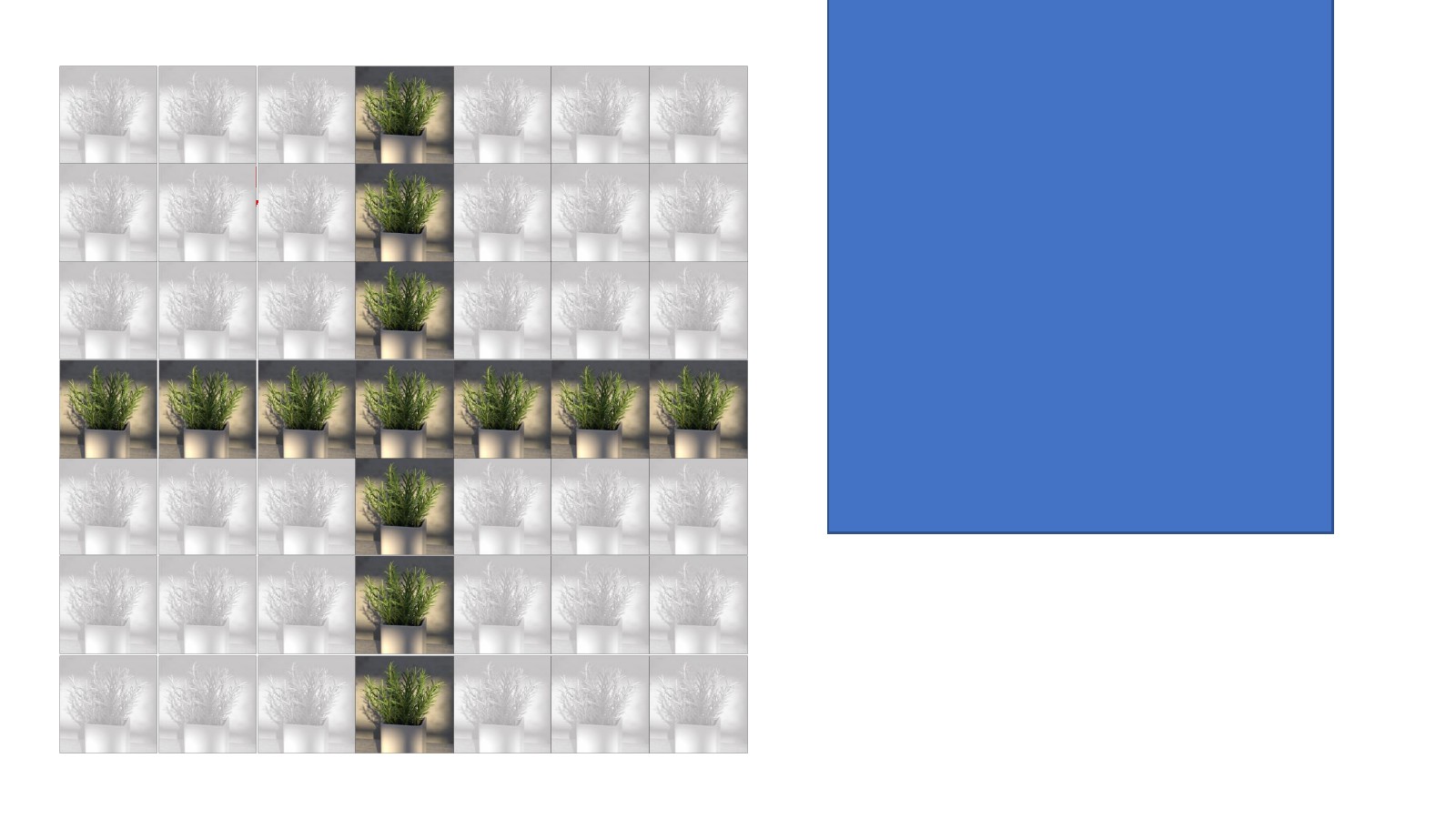} 
			\end{minipage}}
			& 
			\noindent\makebox[.09\textwidth][c]{
				\begin{minipage}{.09\textwidth}
					\centering
					\rule{0pt}{0ex} 
					\includegraphics[page=1,width=1\textwidth,trim={0cm 0.8cm 16.5cm 1cm},clip]{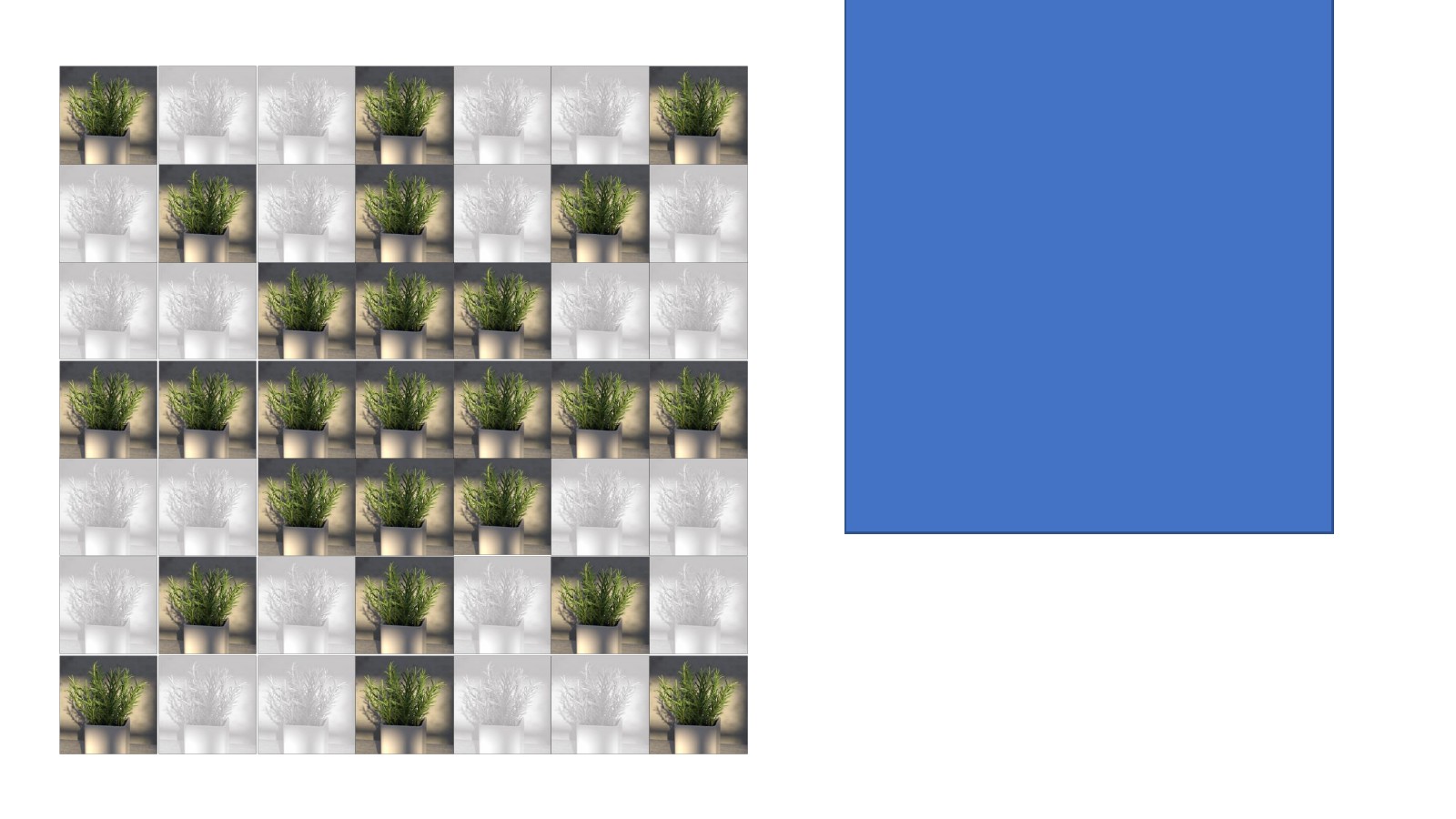}
			\end{minipage}} \\ \hline 
			
			\small{MSE} & \small{ 2.165} & \small { 1.729}  & \small { \bf1.393} \\ \hline
			\small{Bad pixel ratio} & \multirow{2}{*}{\small { 7.61}} & \multirow{2}{*}{\small { 5.94}} & \multirow{2}{*}{\bf \small { 3.87}} \\
			(\small{$<$0.07px)}    &                       &                       &                       \\    \hline
			
		\end{tabular}
	\end{table}			

	To show the effect of different number of streams in the network, we compare the performance of our network with varying number of the streams.
	Using the same architecture with almost the same number of parameters (5.1M), we compare the difference in performance of our network with different numbers of streams in \tref{tab:stream}.
	The network with four streams shows the best performance in terms of the bad pixel ratio and the mean square error. 
	
	After the multi-stream part, we concatenate all the features from each stream, and the size of the feature becomes four times larger. The merge network consists of eight convolutional blocks that finds the relationships between the features passed through the multi-stream network. The blocks in the merge network have the same convolutional structure with that of the multi-stream network except for the last block. 
	To inferring the disparity values with sub-pixel precision, we construct the last block with a ‘Conv-ReLU-Conv’ structure.
	
	\vspace{0.2cm} 
	
	\subsection{Data Augmentation}
	Although there are some public light-field image datasets, only a few of them have data that are under similar conditions as the real light field image with the ground-truth disparity maps.
	In this paper, we use the 16 light-field synthetic images containing various textures, materials, objects and narrow baselines provided in~\cite{Honauer16}. 
	However, 16 light-field images are just not enough to generalize convolutional neural networks.
	To prevent the overfitting problem, data augmentation techniques are essential.
	Therefore, we propose a light-field image specific data augmentation technique that preserves the geometric relationship between the sub-aperture images.
	
\begin{figure}[t]
		\centering
		\hspace{0em}
		\includegraphics[page=1,width=0.45\linewidth,trim={0cm 0cm 14cm 0.1cm},clip]{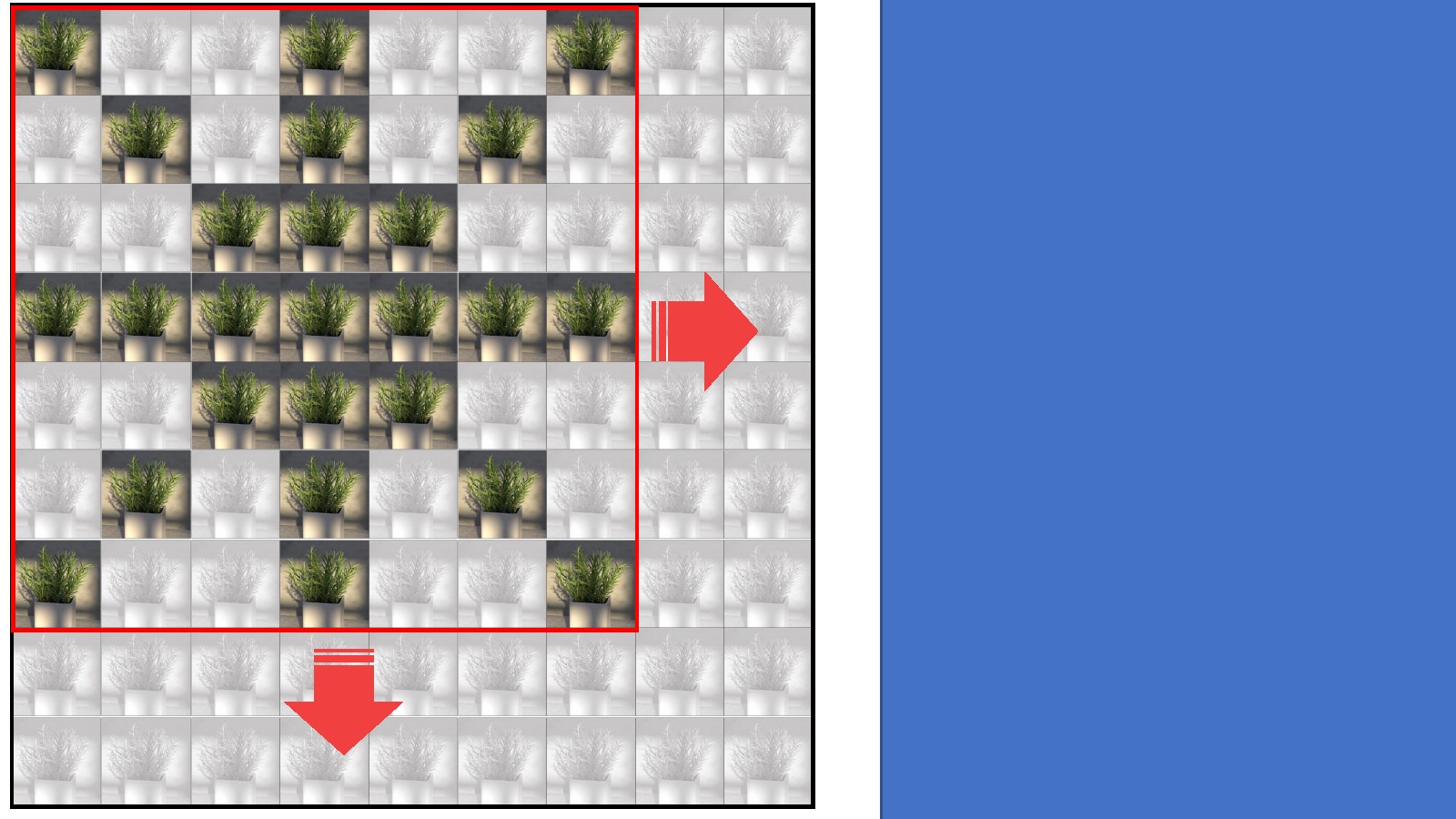}
		\caption{An example of the viewpoint augmentation: (-1,-1) shift.}
		\label{Fig:f3}
		\vspace{-0.2cm}
	\end{figure}

	Our first strategy is to shift the center view point of the light field image.
	The synthetic light field dataset that we used has 9$\times$9 views, each with a 512$\times$512 spatial resolution.
	As shown in~\fref{Fig:f3}, we select the 7$\times$7 views and disparity image of its center view to train our network. 
	By shifting the center view, we can get nine times more training sets through this view-shifting strategy. 
	\begin{figure}[t]
		\vspace{-0.2cm}
		\hspace{0.1cm}
		\includegraphics[page=1,width=0.97\linewidth, trim={0cm 1.1cm 15.3cm 0.2cm},clip]{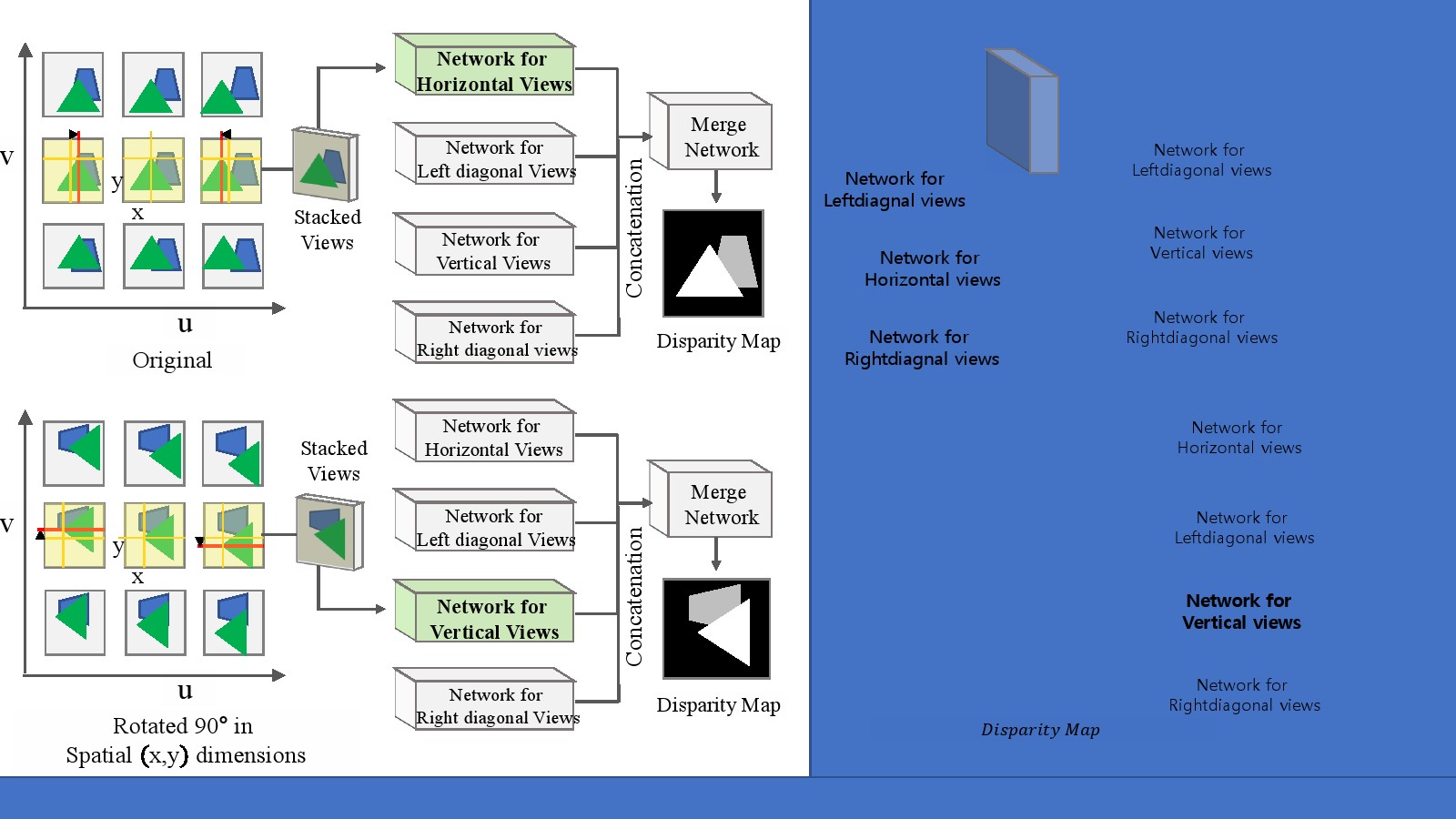}
		\caption{Data augmentation with rotation. When augmenting the data with rotation, we also need to rearrange the connections to put the stacked views into the
			correct network stream.}
		\label{Fig:f4}
		\vspace{-0.2cm}
	\end{figure}	
	To validate the performance according to the number of viewpoints and the view-shifting augmentation, we compare performances of the networks using 3$\times$3, 5$\times$5, 7$\times$7, and 9$\times$9 input views.
	As shown in~\tref{tab:AngularAug}, we found that there are performance gains when increasing the number of input views. 
	However, the gain is marginal when comparing  9$\times$9 views with 7$\times$7 views.
	The 7$\times$7 views shows the better performance in the mean square error.
	This shows the effectiveness of our view-shifting augmentation. 

	We also propose a rotation augmentation method for light field images. As in the depth estimation~\cite{mayer2016large} and the optical flow estimation~\cite{dosovitskiy2015flownet} using deep learning, the image rotation in the spatial dimension has been widely used as an augmentation technique.
	However, the conventional rotational method cannot be directly used, as it does not consider the directional characteristics of the light-field image. 
	\begin{table*}[t]
		\centering
		\caption{Effects of the angular resolutions and the augmentation techniques on performance.}
		\label{tab:AngularAug}
		\begin{tabular}{|c|c|c|c|c|c|c|c|c|}
			\hline
			\begin{tabular}[c]{@{}c@{}}\small{Augular} \vspace{-0.1cm} \\     \small{resolution} \end{tabular}              & \small{3 $\times$ 3}      & \small{5 $\times$ 5}      & \multicolumn{5}{c|}{\small{7 $\times$ 7}} & \small{9 $\times$ 9}   \\ \hline
			\begin{tabular}[c]{@{}c@{}}\small{Augmentaion type}\end{tabular}                  & \small{Full Aug} & \small{Full Aug} & \small{Color} & \begin{tabular}[c]{@{}c@{}}\small{Color +} \vspace{-0.1cm}\\     \small{Viewshift}\end{tabular} & \begin{tabular}[c]{@{}c@{}}\small{Color +} \vspace{-0.1cm}\\     \small{Rotation}   \end{tabular} & \begin{tabular}[c]{@{}c@{}}\small{Color +} \vspace{-0.1cm}\\     \small{scaling}\end{tabular} & \small{Full Aug} & \small{Full Aug} \\ \hline \hline
			\small{Mean square error} & \small{1.568}    & \small{1.475}    & \small{2.799} & \small{2.564}   & \small{1.685}   & \small{2.33}   & \small{1.434}    & \small{1.461}    \\ \hline 
			\begin{tabular}[c]{@{}c@{}}\small{Bad pixel ratio} \vspace{-0.1cm}\\ \small{(\textgreater0.07px)} \end{tabular} & \small{8.63 }    & \small{4.96}     & \small{6.67}  & \small{6.29} & \small{5.54}  & \small{5.69}   & \small{3.94 }    & \small{3.91}     \\ \hline
		\end{tabular}
		\vspace{-0.1cm}
	\end{table*}  			
	In the multi-stream part of our network, we extract features for epipolar property of viewpoint sets.
	To preserve this light field property, we first rotate sub-aperture images in the spatial dimension, and then rearrange the connection of the viewpoint sets and the streams as shown in \fref{Fig:f4}.
	This rearrangement is necessary as the geometric properties change after the rotation.
	For example, pixels in the vertical direction are strongly related to each other in the vertical views as mentioned in Section 3.1.
	If we rotate the sub-aperture images in the vertical viewpoints by 90 degrees, it makes the horizontal view network stream to look at vertical characteristics.
	Thus, the rotated sub-aperture images should be input to the vertical view stream.

	We additionally use general augmentation techniques such as the scaling and the flipping. 
	When the images are scales, the disparity values need to be also scaled accordingly. 
	We tune the scales of both the image and the disparity by $1/N$ times $(N=1,2,3,4)$.
	The sign of disparity is reversed when flipping the light field images.
	With these augmentation techniques: the view-shifting, rotation [90, 180, 270 degrees], image scaling [0.25, 1], color scaling [0.5, 2], randomly converting color to gray scale from [0, 1]; gamma value from [0.8, 1.2] and flipping, we can increase the training data up to 288 times the original data.
	
	We validate the effectiveness of the light-field specific augmentations.
	As seen in~\tref{tab:AngularAug}, there are large performance gains when using the rotation and flipping. 
	We also observe that the scaling augmentation allows to cover various disparity ranges, which is useful for real light-field images with very narrow baselines. 
	Through the augmentation techniques, we reduce disparity errors by more than 40$\%$.

	\subsection{Details of learning}
\begin{figure}[t]
		\hspace{0em}
		\begin{tabularx}{1.0\linewidth}{@{\hspace{0cm}}c@{\hspace{0.1cm}}c@{\hspace{0.1cm}}c@{\hspace{0.1cm}}c}
			& \footnotesize{Kitchen}  & \footnotesize{Museum} & \footnotesize{Vinyl} \vspace{0cm}\\
			\rotatebox[origin=l]{90}{\footnotesize{\hspace{0.5cm}Failure case}}&
			{\includegraphics[ width=2.55cm, height=2.55cm]{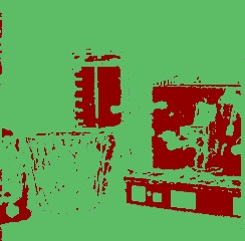}} &
			\includegraphics[ width=2.55cm, height=2.55cm]{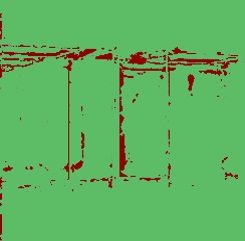} &
			\includegraphics[ width=2.55cm, height=2.55cm]{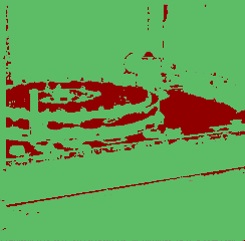} \\
			\rotatebox[origin=l]{90}{\footnotesize{\hspace{0.28cm}Reflection mask}}&
			{\includegraphics[ width=2.55cm, height=2.55cm]{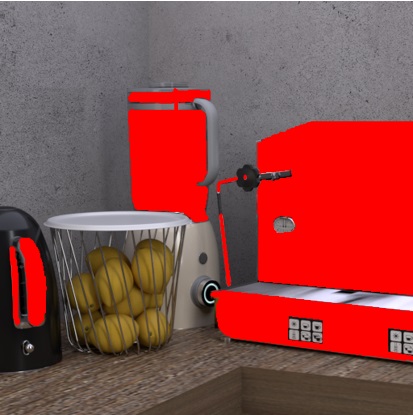}} &
			\includegraphics[ width=2.55cm, height=2.55cm]{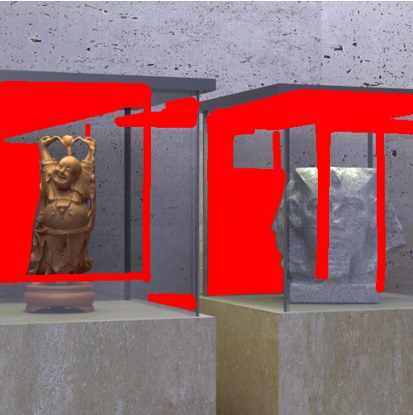} &
			\includegraphics[ width=2.55cm, height=2.55cm]{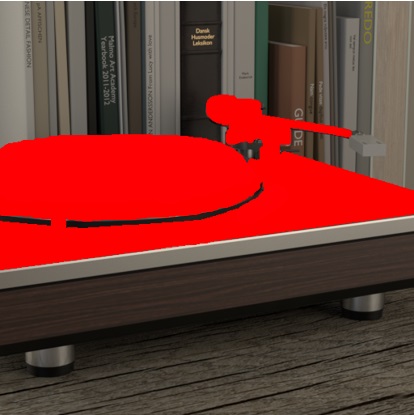} \vspace{0.1cm}\\
		\end{tabularx}		
		\caption{(Top) Failure cases in regions with reflections. (Bottom) Examples of reflection masks for training data.}
		\label{Fig:f8}
		\vspace{-0.4cm}
	\end{figure}

	We exploit the patch-wise training by randomly sampling gray-scale patches of size 23$\times$23 from the 16 synthetic light-field images~\cite{Honauer16}.
	To boost training speed, all convolutions in the layers are conducted without zero padding.
	We exclude some training data that contains reflection and refraction regions such as glass, metal and textureless regions, which result in incorrect correspondences. 
	Reflection and refraction regions were manually masked out in ~\fref{Fig:f8}.
	We also removed textureless regions where the mean absolute difference between a center pixel and other pixel in a patch is less than 0.02. 
	
	As the loss function in our network, we used the mean absolute error (MAE) which is robust to outliers~\cite{goodfellow2016deep}.
	We use Rmsprop ~\cite{tieleman2012lecture} optimizer and set the batch size to 16. The learning rate started at 1e-5 and is decreased to 1e-6. 
	Our network takes 5$\sim$6 days to train on a NVIDIA GTX 1080TI and is implemented in TensorFlow~\cite{tensorflow2015-whitepaper}.
	\vspace{0.05cm}
	\section{Experiments}
	In this section, the performance of the proposed algorithm is evaluated using synthetic and real-world datasets.
	The 4D light field benchmark ~\cite{Honauer16} was used for the synthetic experiments.
	The benchmark has 9$\times$9 angular and 512$\times$512 spatial resolutions.
	For real-world experiments, we utilized images captured with a Lytro illum~\cite{LytroIllum}.

	\subsection{Quantitative Evaluation}\label{sec:quant_eval}
	For the quantitative evaluation, we estimate the disparity maps using the test sets in the 4D Light Field Benchmark~\cite{Honauer16}.
	Bad pixel ratios and mean square errors were computed for the 12 light-field test images. 
	Three thresholds (0.01, 0.03 and 0.07 pixels) for the bad pixel ratio are used, in order to better assess the performance of algorithms for difficult scenes. 
	
	\begin{figure*}[t]
		\centering
		\begin{tabularx}{1.0\linewidth}{@{\hspace{1.9cm}}c@{\hspace{2.8cm}}c@{\hspace{3.0cm}}c@{\hspace{1.25cm}}c}
			\footnotesize{Bad pixel} & \footnotesize{Bad pixel}  & \footnotesize{Bad pixel}  & \footnotesize{Mean Square Error}  \vspace{-0.1cm}\\
			\footnotesize{(Error$<$0.01)}& \footnotesize{(Error$<$0.03)}  & \footnotesize{(Error$<$0.07)} & \footnotesize{(multiplied with 100)} \\
			
			\multicolumn{4}{c}{	\includegraphics[page=1,width=1.03\linewidth,trim={0cm 8.5cm 0cm 1.3cm},clip]{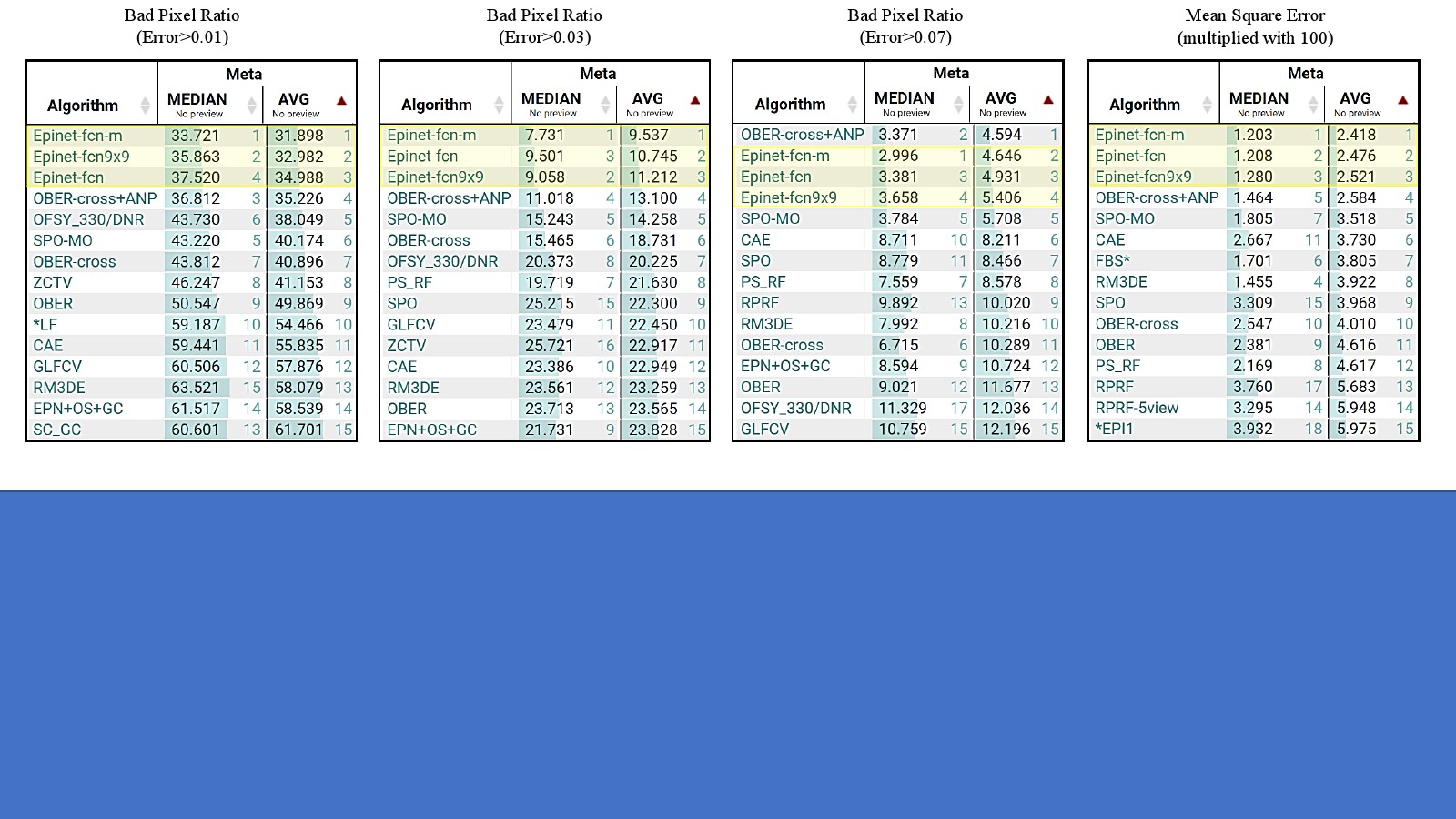}} \\
			
		\end{tabularx}
		\caption{Benchmark ranking (http://hci-lightfield.iwr.uni-heidelberg.de). Several versions of the proposed methods are highlighted.  }
		\label{Fig:f5}
	\end{figure*}
	
	\begin{figure}[t]
		\hspace{0.4cm}
		\begin{tabularx}{1.00\linewidth}{@{\hspace{0cm}}c@{\hspace{0cm}}}
			\small{Runtime (seconds)} \vspace{-0.0cm}\\		
			\includegraphics[page=1,width=1.0\linewidth, trim={0cm 10cm 17cm 0.3cm},clip]{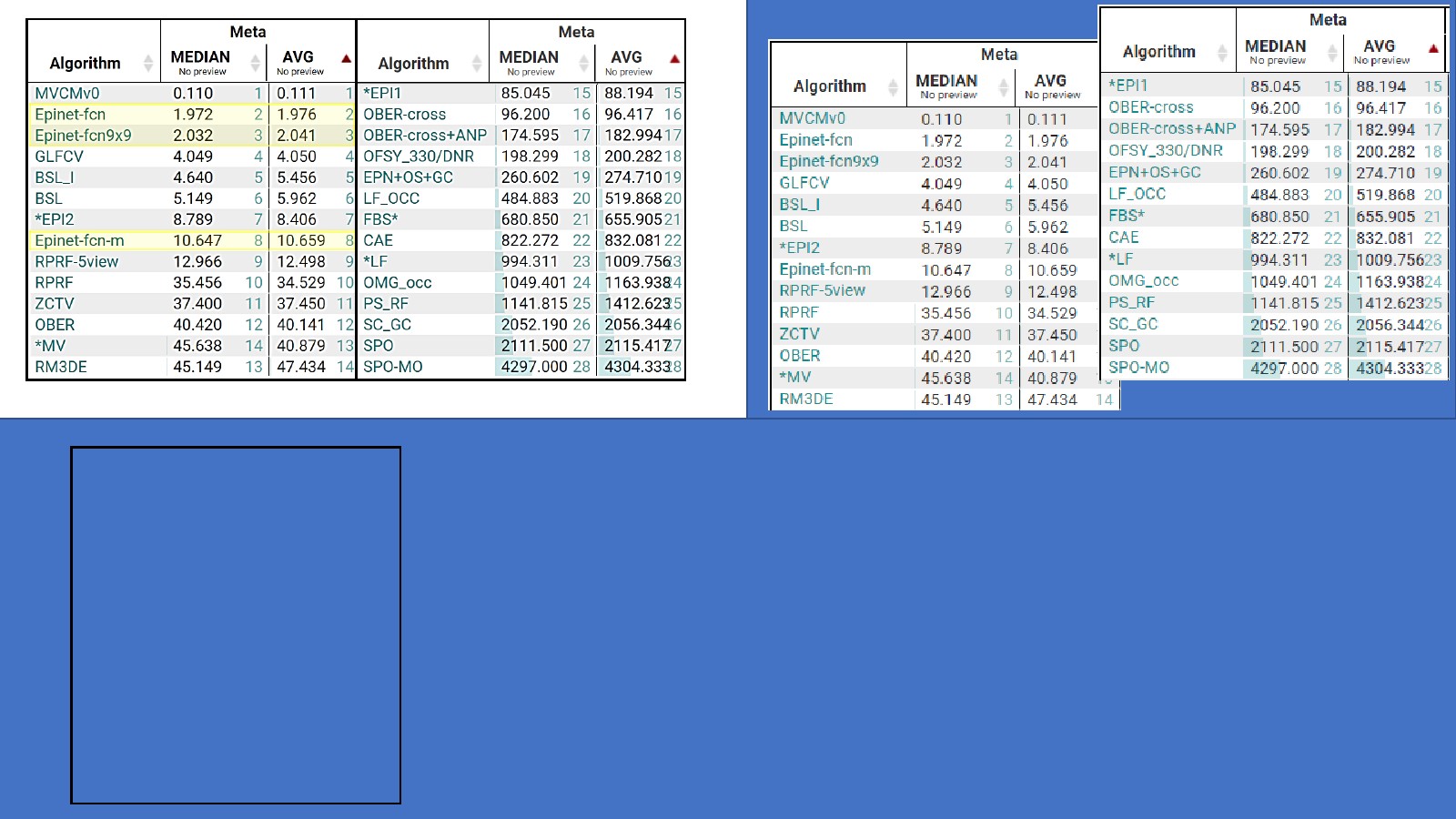}
		\end{tabularx}
		\caption{Runtime benchmark of the algorithms}
		\label{Fig:runtime} \vspace{0.2cm}
	\end{figure}
	
	In~\fref{Fig:f5}, we directly refer to the ranking tables, which are published on the benchmark website. 
	Our EPINET shows the best performance in 3 out of 4 measures. 
	Epinet-fcn is our EPINET model using the vertical, the horizontal, the left diagonal and the right diagonal viewspoints as input, and Epinet-fcn9x9 is a model that uses all 9$\times$9 viewpoints. The Epinet-fcn-m is a modified version of our Epinet-fcn. Epinet-fcn-m predicts multiple disparity maps by flipping and rotating (90, 180, 270 degrees) the given light field image. The final estimation is the average of the estimated disparity maps, which reduces the matching ambiguity.
	In addition to the accuracy, the EPINET is effectively the fastest algorithm among the state-of-the-art methods
	as shown in ~\fref{Fig:runtime}.
	Our computational time is second to MVCMv0, but its depth accuracy is the last in the benchmark. 

\begin{table}[t]
\footnotesize
\centering
\caption{ Quantitative evaluation of deep learning based methods using 50 synthetic LF images. The table provides the RMSE and MAE. For \cite{Heber16, heber2017neural}, the error metrics are directly referred from \cite{heber2017neural}.}
\label{tab:quan}
\begin{tabular}{lclclclc}
    Method (\# of training images)    & RMSE & MAE   & Time \\\hline
{[}12{]} (850) & 1.87 & 1.13 & 35s  \\
{[}13{]} (850) & 0.83 & 0.34 & 0.8s \\
Ours (250)     &   0.14   &   0.036    &   0.93s  
\end{tabular}
\end{table}
	\begin{figure}[t]
		\centering
		\includegraphics[page=1,width=0.48\textwidth,trim={0cm 12cm 7cm 0cm},clip]{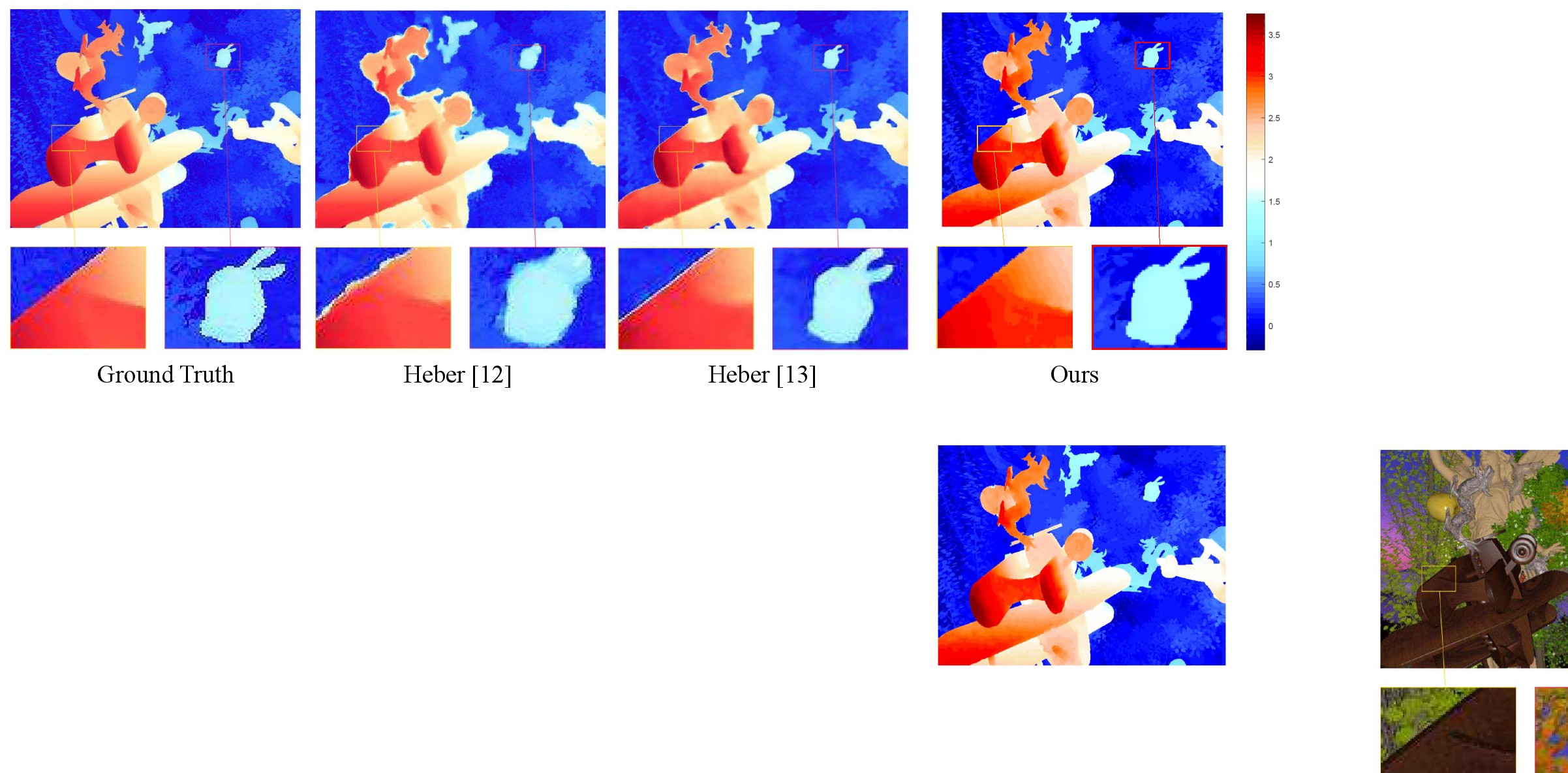}
		\caption{Comparison with deep learning-based methods \cite{Heber16, heber2017neural}. The results for \cite{Heber16, heber2017neural} are directly referred from \cite{heber2017neural}.}
		\label{Fig:compare_heber} \vspace{-0.3cm}
	\end{figure}	

\begin{figure*}[t]
		\includegraphics[page=1,width=1.01\linewidth,trim={0.5cm 0.97cm 2.5cm 0cm},clip]{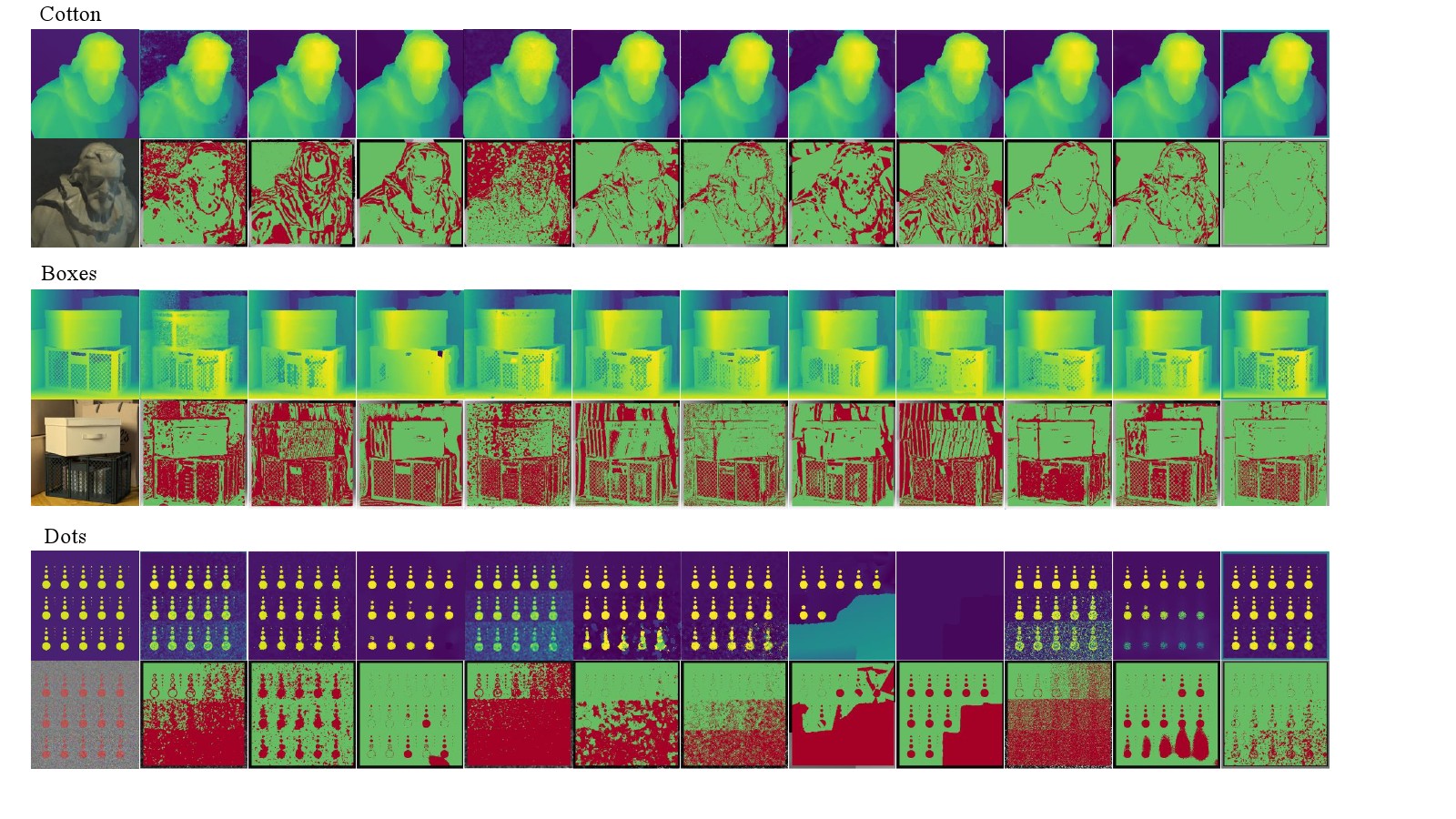}
		\\
		\begin{tabularx}{1.01\textwidth}{@{\hspace{-0.1cm}}p{1.32cm}@{\hspace{0cm}}p{1.44cm}@{\hspace{0cm}}p{1.44cm}@{\hspace{0cm}}p{1.44cm}@{\hspace{0cm}}p{1.44cm}@{\hspace{0cm}}p{1.44cm}@{\hspace{0cm}}p{1.44cm}@{\hspace{0cm}}p{1.44cm}@{\hspace{0cm}}p{1.44cm}@{\hspace{0cm}}p{1.44cm}@{\hspace{0cm}}p{1.44cm}@{\hspace{0cm}}p{1.44cm}}
			\hspace{0.31cm}(a) GT &
			\hspace{0.34cm}(b)~\cite{Wanner12} &
			\hspace{0.34cm}(c)~\cite{wang2015occlusion}&
			\hspace{0.34cm}(d)~\cite{Jeon15} &
			\hspace{0.34cm}(e)~\cite{johannsen2016sparse}&
			\hspace{0.34cm}(f)~\cite{Williem16} &
			\hspace{0.34cm}(g)~\cite{Zhang16} &
			\hspace{0.34cm}(h)~\cite{Lipeng2016Dense} &
			\hspace{0.34cm}(i)~\cite{huang2017robust} &
			\hspace{0.34cm}(j)~\cite{SAG17cvpr} &
			\hspace{0.34cm}(k)~\cite{PS_RF} &
			\hspace{0.34cm}(l) Ours \vspace{-0.3cm} \\ 
		\end{tabularx}
		\caption{Qualitative results of the HCI light-field benchmark. Odd rows shows the estimated disparity results and even rows represent error maps for bad pixel ratio of 0.03.}
		\label{Fig:f6}		\vspace{-0.5cm}
	\end{figure*}

	Qualitative results (\textit{Cotton, Boxes and Dots}) are shown in~\fref{Fig:f6}. 
	The \textit{Cotton} scene contains smooth surfaces, and the \textit{Boxes} scene is composed of slanted objects with depth discontinuity occlusions.
	As can be seen from the examples, our EPINET reconstructs the smooth surface and the sharp depth discontinuity better than previous methods. 
	The EPINET infers accurate disparity values through the regression part in the network as our fully-convolutional layers can precisely distinguish the subtle difference of EPI slopes.
	The \textit{Dots} scene suffers from image noise whose levels varies spatially.
	Once again, the proposed method achieves the best performance in this noisy scene because the 2$\times$2 kernel has the effect of alleviating the noise effect.

	     A direct comparison between the EPINET and other state-of-the-art deep learning-based approaches \cite{Heber16, heber2017neural} can be found in~\tref{tab:quan} and \fref{Fig:compare_heber}.
         We trained the EPINET on 250 LF images provided by the authors of \cite{Heber16, heber2017neural} whose baseline is (-25, 5) pixels.
		The EPINET still outperforms the works in~\cite{Heber16,heber2017neural}.
		Our multi-streams strategy to resolve the directional matching ambiguities enables to capture sharp object boundaries like the airplane's wing and the toy's head.
		Another reason for the better performance is that the LF images of \cite{Heber16, heber2017neural} contain highly textured regions with less noise compared to the HCI dataset.

	\subsection{Real-world results}
	
	We demonstrate that our EPINET also achieves reliable results on real light-field images.
	We used the light-field images captured by a Lytro illum camera~\cite{LytroIllum}, provided by the authors of ~\cite{Bok17}.
	The real-world dataset is challenging as the data contain a combination of smooth and slanted surfaces with depth discontinuity. 
	Additionally, these images suffer from severe image noise due to the inherent structural problem in the camera. 
	\begin{figure}[t]
    	\centering
    	\hspace{0em}
    	\includegraphics[page=1,width=0.75\linewidth, trim={0cm 2.6cm 0cm 0cm},clip]{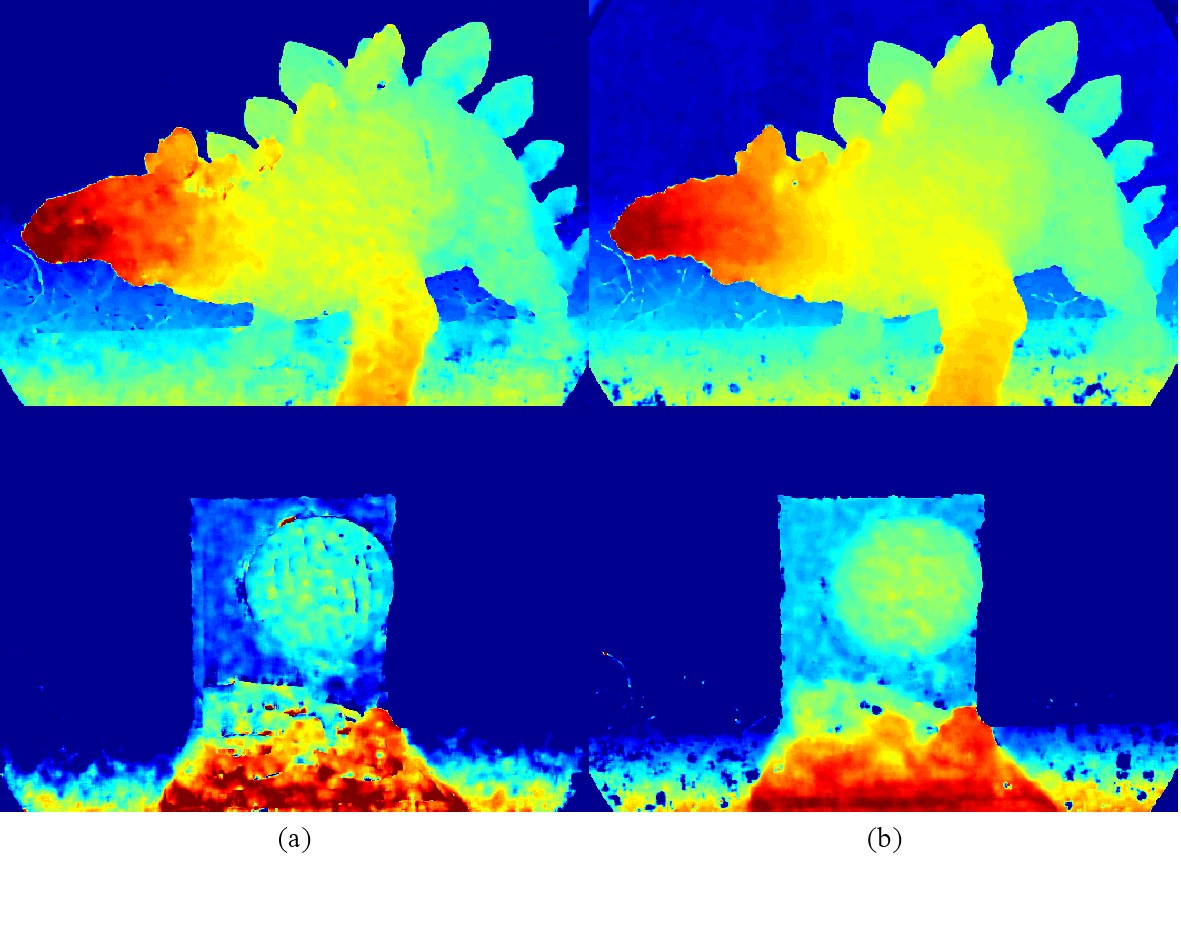}
    	\\ \hspace{0.05cm}  \small{ \hspace{-0.2cm} (a) 7$\times$7 \hspace{2.5cm} (b) 9$\times$9} 
    	\caption{Real world results using 7x7 angular resolution and 9x9 angular resolution of light field images.} \vspace{-0.4cm}
    	\label{Fig:7x7_9x9} \vspace{-0.4cm}
	\end{figure}
	In~\fref{Fig:7x7_9x9}, we compare the disparity predictions from the EPINET using the input viewpoints 7$\times$7 and 9$\times$9. 
	Although the performances of both EPINETs are similar in the synthetic data, there is a noticeable performance differences between the two in the real-world. 
	In \cite{Wanner13}, it has been shown that the accuracy of the depth estimation from light field improves with more viewpoints, since they represent a consensus of all input views.
	Thus, we used the EPINET with 9$\times$9 input viewpoints for the real-world experiments.
	We also remove sparse disparity errors using the conventional weighted median filter~\cite{Ma13} for only the real-world dataset.
	\begin{figure}[t]
		\centering
		\includegraphics[page=1,width=0.8\linewidth, trim={0cm 0cm 0cm 0cm},clip]{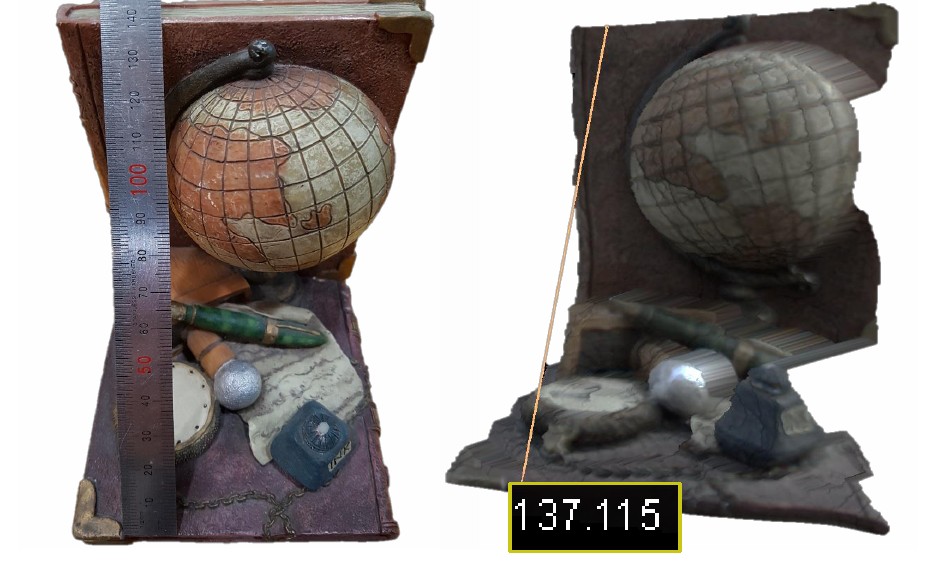} 
		\caption{Mesh rendering result of our disparity. (unit: mm)}
		\label{Fig:mesh} \vspace{-0.4cm}
	\end{figure}
	
\begin{figure*}[t]
        \centering 
    	\includegraphics[width=0.88\linewidth,trim={2cm 9.57cm 1cm 0cm},clip]{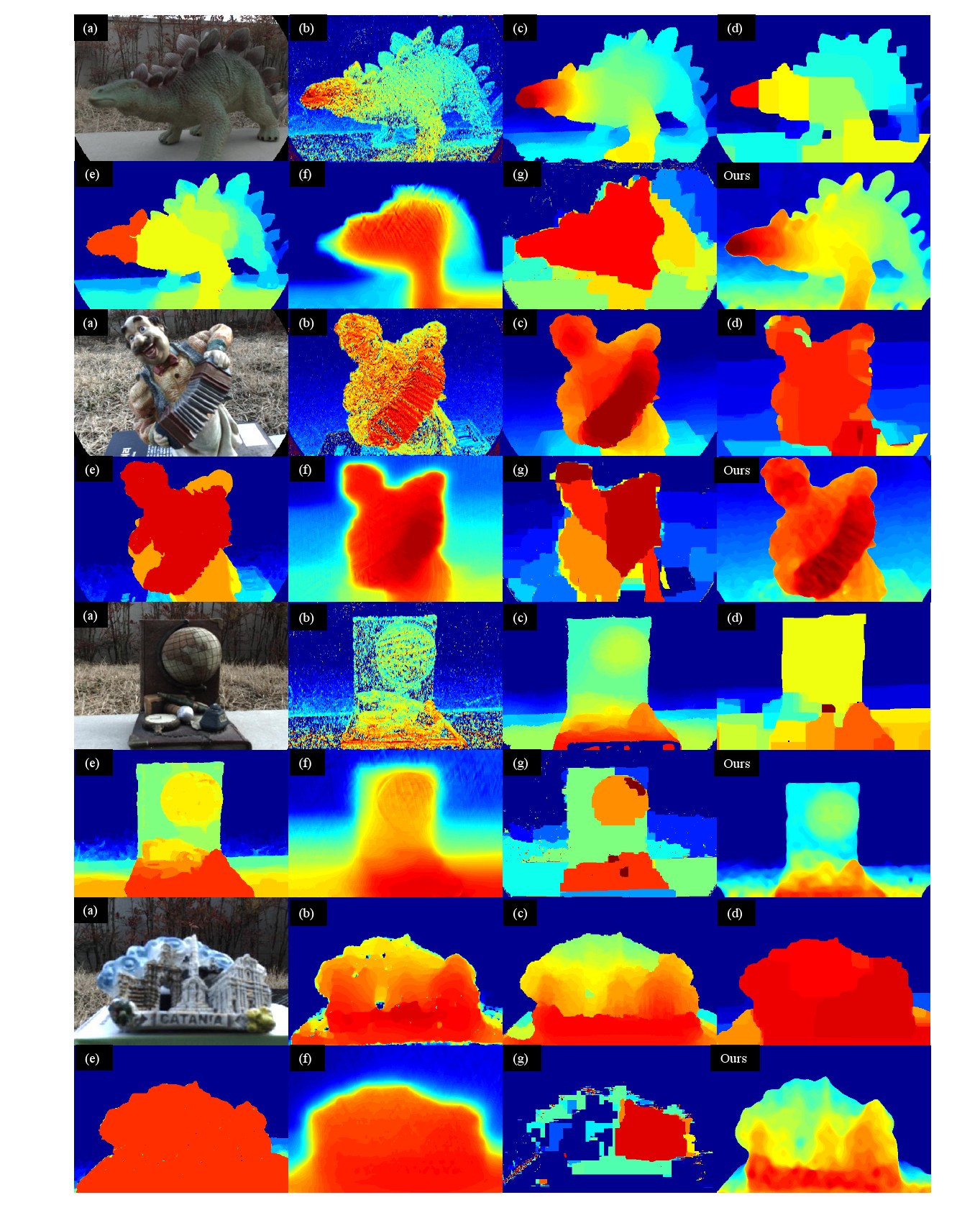}		
    	\caption{Qualitative results of real world data. (a) reference view (center view), (b)~\cite{Wanner12}, (c)~\cite{Jeon15}, (d)~\cite{Williem16}, (e)~\cite{wang2015occlusion}, (f)~\cite{Tao16}, (g)~\cite{Yu13}}
    	\label{Fig:f7}
    \end{figure*}

	\fref{Fig:f7} compares qualitative results with previous methods. 
	Although the work in~\cite{Jeon15} shows good results, the method requires several minutes for the depth estimation with careful parameter tuning.
	In contrast, our EPINET achieves the state-of-the-art results much faster, without any parameter tuning. 
	An accurate disparity map can facilitate many applications. 
	As an example, we reconstructed the 3D structure of an object captured by the Lytro illum camera in the metric scale using the depth computed with our method.
	In~\fref{Fig:mesh}, the disparity map was converted into a 3D model in the metric scale using the calibration parameters estimated by the toolbox of~\cite{Bok17}.
	The impressive results show that the proposed method can be used for further applications like the 3D printing. 

	\section{Conclusion}

	In this paper, we have proposed a fast and accurate depth estimation network using the light field geometry.

	Our network has been designed taking into account the light field epipolar geometry to learn the angular and the spatial information using a combination of a multi-stream network and a merging network.
	In addition, we introduced light-field image-specific data augmentations such as view-shifting and rotation.
	Using the proposed method, we could overcome insufficient data problem and show the state-of-the-art results on the Benchmark light-field images as well as real-world light-field images.

	There are still rooms for improving our method. 
	First, the easiest way to improve the CNN-based approach is to boost the number of realistic dataset.
	Second, our network fails to infer accurate disparities in reflection and textureless regions.
	To handle this issue, we think that a prior knowledge such as object material~\cite{Wang16material} can be included in the future work. 
	We also expect that our network model can be improved by fusing a photometric cue~\cite{Tao16} or a depth boundary cue~\cite{Wang16}. 
	\section*{Acknowledgement}
	This work was supported by the National Research Foundation
of Korea (NRF) grant funded by the Korea government
(MSIP) (NRF-2016R1A2B4014610).
	\clearpage
	{\small
		\bibliographystyle{ieee}
		\bibliography{egbib}
	}
	
\end{document}